%% file: formatting-instructions-latex-2020.tex
\def\year{2021}\relax
\documentclass[letterpaper]{article} 
\usepackage{aaai21}  
\usepackage{times}  
\usepackage{helvet} 
\usepackage{courier}  
\usepackage[hyphens]{url}  
\usepackage{graphicx} 
\usepackage{graphicx}  
\usepackage{mathtools}
 
\usepackage{dsfont}
\usepackage{enumitem}
\usepackage{caption}
\usepackage{multirow}
\usepackage{amsmath,amssymb} 

\usepackage{amsthm}
\theoremstyle{plain}

\usepackage[section]{placeins}

\usepackage{epsfig}
\usepackage{amsmath,amssymb} 
\usepackage{multirow}

\usepackage{algorithm}
\usepackage{algorithmic}

\title{Subtype-aware Unsupervised Domain Adaptation \\ for Medical Diagnosis}

\author{Xiaofeng Liu{$^{1,2}$},~Xiongchang Liu{$^{2,3\dag}$},~Bo Hu{$^{2,4\dag}$},~Wenxuan Ji{$^{5}$},~Fangxu Xing{$^{1}$},~Jun Lu{$^{2*}$},\\
{\Large\textbf{Jane You{$^{6}$},~C.-C. Jay Kuo{$^{7}$},~Georges El Fakhri{$^{1}$},~Jonghye Woo{$^{1}$}}}\\
{$^{1}$}Dept. of Radiology, Massachusetts General Hospital and Harvard Medical School, Boston, MA, USA\\{$^{2}$}Dept. of Neurology, Beth Israel Deaconess Medical Center and Harvard Medical School, Boston, MA, USA\\{$^{3}$}China University of Mining and Technology, China {$^{4}$}Beijing University of Posts and Telecommunications, China  \\{$^{5}$}School of Artificial Intelligence, Nankai University, China \\{$^{6}$}Dept. of Computing, The Hong Kong Polytechnic University, Hong Kong\\{$^{7}$}Dept. of Electrical and Computer Engineering, University of Southern California, Los Angeles, CA, USA\\
{\small{{$^{\dag}$}Co-second authors contribute equally~~{$^{*}$}Corresponding Author}}
}

\begin{document}

\maketitle

\input{0_Abstract.tex}

\input{1_Introduction.tex}

\input{2_RelatedWork.tex}

\input{3_Approach.tex}

\input{4_Experiments.tex}

\input{5_Conclusion.tex}

\bibliographystyle{aaai} \bibliography{egbib}

\end{document}

%% file: 0_Abstract.tex
\begin{abstract}

Recent advances in unsupervised domain adaptation (UDA) show that transferable prototypical learning presents a powerful means for class conditional alignment, which encourages the closeness of cross-domain class centroids. However, the cross-domain inner-class compactness and the underlying fine-grained subtype structure remained largely underexplored. In this work, we propose to adaptively carry out the fine-grained subtype-aware alignment by explicitly enforcing the class-wise separation and subtype-wise compactness with intermediate pseudo labels. Our key insight is that the unlabeled subtypes of a class can be divergent to one another with different conditional and label shifts, while inheriting the local proximity within a subtype. The cases with or without the prior information on subtype numbers are investigated to discover the underlying subtype structure in an online fashion. The proposed subtype-aware dynamic UDA achieves promising results on a medical diagnosis task. 

\end{abstract}

%% file: 1_Introduction.tex
\section{Introduction}

The goal of unsupervised domain adaptation (UDA) is to transfer knowledge learned from a label-rich domain to new unlabeled target domains \cite{saito2017adversarial,liu2020energy,liu2020disentanglement,zou2019confidence}. The conventional adversarial training and maximum mean discrepancy (MMD) based methods propose to align the marginal distribution of sample $x$ in the feature space, i.e., $p(f(x))$, where $f(\cdot)$ is a feature extractor. Given the Bayes' theorem $p(f(x)|y)=\frac{p(y|f(x))p(f(x))}{p(y)}$, suppose that there are no concept and label shifts (i.e., $p(y|f(x))$ and $p(y)$ are the same for two domains), then the conditional distribution $p(f(x)|y)$ can be aligned by aligning $p(f(x))$. Nonetheless, the label shift $p(y)$ is quite common in real-world applications, which indicates the label proportion is different \cite{zhao2019learning}.

 
 
\begin{figure}[t]
\centering
\includegraphics[width=8.5cm]{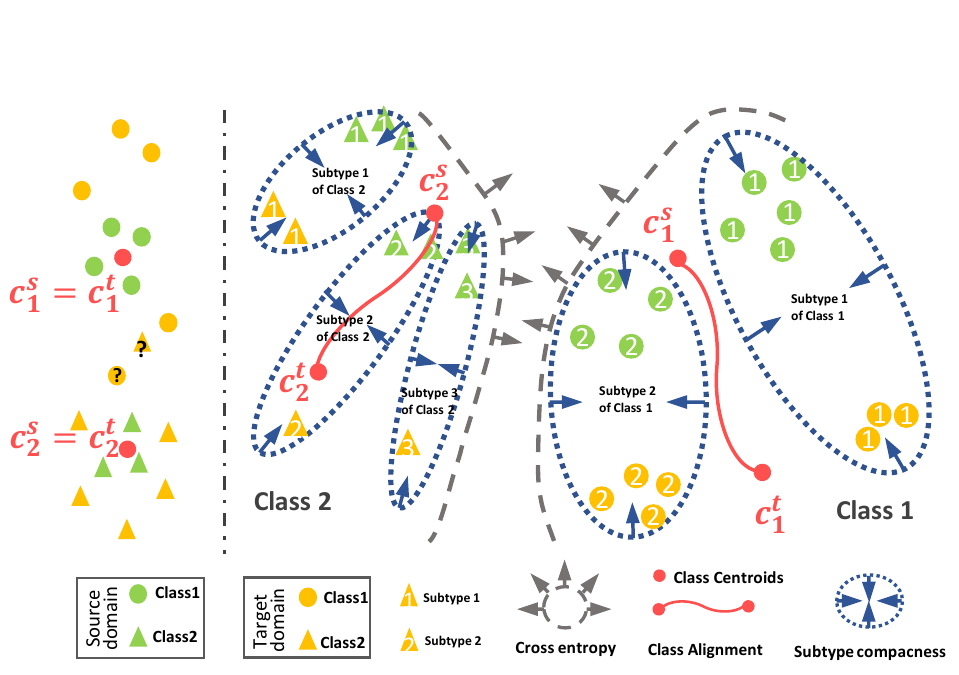}\\
\caption{Illustration of the failure case of prototypical UDA (left), and the idea of our subtype-aware UDA (right).}\label{fig:11} 
\end{figure}




Recently, transferable prototypical networks (TPN) \cite{pan2019transferrable} is proposed to promote the source domain class separation with a cross-entropy (CE) loss, and match the class centroids of source and target samples to perform class-wise alignment. However, the CE loss in the source domain cannot minimize the inner-class variation \cite{liu2016large}. The class-wise separation in the to-be tested target domain cannot be well supported by the centroid closeness objective. As shown in Fig. \ref{fig:11} left, although the class centroids are well aligned, the sparsely distributed target samples can be easily misclassified. 
 
One way to tackle this is by simply enforcing the cross-domain inner-class feature distribution compactness. However, in many cases, the unlabeled subtypes in a class can be diverse, and form an underlying local distribution. For instance, different cancer subtypes may have significantly diverse patterns \cite{yeoh2002classification}. In such circumstances, the shared pattern among two different subtypes may not be exclusive for class-level discrimination. In these applications, it would be more reasonable and effective to exploit the subtype-wise patterns. Moreover, unsupervised deep clustering \cite{caron2018deep} empirically assigns 10$\times$ more clusters of the class to contain diverse subtypes. Recent works also show the fine-grained label can be helpful for the coarse classification \cite{chen2019understanding}.


Moreover, domain shifts can be different w.r.t. subtypes, which leads to \textit{subtype conditional shift}. Besides, the incidence of disease subtypes is usually varied across different regions, which leads to \textit{subtype label shift}. The proportion difference at the subtype-level can usually be more significant than the class-level \cite{wu2019domain}. This motivates us to extend the concept of class conditional and label shifts \cite{kouw2018introduction} to the fine-grained subtype-level (i.e., $p(f(x)|k)$ and $p(k)$ vary across domains for subtype $k$). Therefore, a more realistic presumption of UDA can be both the class and subtype conditional and label shifts.



In this work, we resort to the feature space metric learning with intermediate pseudo labels to adaptively achieve both class-wise separation and cross-domain subtype-wise compactness. We first propose an online clustering scheme to explore the underlying subtype structure in an unsupervised fashion. With the prior knowledge of subtype numbers, concise $k$-means clustering can be simply applied, by assigning $k$ to the subtype numbers. However, the subtype can be challenging to define, due to different taxonomy. We thereby further expand on our framework to unknown subtype numbers by capturing the underlying subtype structure with an adaptive sub-graph scheme using a reliability-path. With a few meta hyperparameters shared between clusters, the sub-graph scheme can be scalable to several classes and subtypes. To explicitly enforce the subtype-wise distribution compactness, the involved samples of a subtype are expected to be close to their subtype centroid in the feature space.



Our main contribution can be summarized as follows:

\begin{itemize}
    \item We propose to adaptively explore the subtype-wise conditional and label shifts in UDA without the subtype labels, and explicitly enforce the subtype-aware compactness.
    
    \item We systematically investigate the cases with or without the prior information on subtype numbers. Our reliability-path based sub-graph scheme can effectively explore the underlying subtype local distribution with a few meta hyperparameters in an online fashion.
    
    \item  We empirically validate its effectiveness on a multi-view congenital heart disease (CHD) diagnosis task with an efficient multi-view processing network and achieve promising performance.

\end{itemize}

%% file: 2_RelatedWork.tex
\section{Related Work}

In recent years, big data drives the fast development of deep learning, which has transformed many fields, such as computer vision and medical image analysis \cite{Han_2020_CVPR_Workshops,liu2019deep,liu2020symmetric}. Deep learning has drastically transformed the way in which features are extracted and then fed into a prediction model into simultaneously learning both features and a prediction model in an end-to-end fashion. The effectiveness of deep learning has been demonstrated in many computer vision tasks, such as classification, detection, and segmentation \cite{Liu_2019_ICCV,liu2020identity,liu2018dependency,liu2019permutation}. In addition, to date, conventional machine learning research in medical image analysis has relied on hand-crafted features \cite{maraci2017framework,liu2018joint,liu2017line} and expert decision rules \cite{de2018clinically}. End-to-end deep learning approaches have also shown promising performance in many disease diagnosis tasks \cite{liu2019unimodal,liu2018ordinal}. For example, \cite{litjens2019state} attempts to analyze ultrasonic data with deep learning and provides a diagnosis suggestion. 




Compared with over one million images of the ImageNet dataset, the collection of large-scale medical data is challenging for clinical applications \cite{liu2020unimodal,liu2018data,He_2020_CVPR_Workshopsb}. To counter this, UDA has gradually become popular \cite{zou2019confidence,liu2020energy}, which aims to match covariate shift (i.e., only $p(x)$ shift). Discrepancy-based methods \cite{long2015learning}, such as minimizing MMD, address the dataset shift by mitigating specific discrepancies defined on different layers of a shared model between domains. Recently, adversarial training utilizes a discriminator to classify the domain to encourage domain confusion \cite{tzeng2017adversarial,liu2020auto3d}. These methods assume that the label proportion is invariant for all of the involved domains \cite{moreno2012unifying}. Yet, the conditional shift \cite{magliacane2018domain} (i.e., only $p(x|y)$ shift) can be more realistic than the covariate shift \cite{zhao2019learning}, and the class label shift (i.e., only $p(y)$ shift) also widely exists in the most of real-world applications \cite{kouw2018introduction}. Furthermore, the subtype-wise conditional and label shifts are a realistic assumption in many applications, which can be more challenging than the class-wise shifts, due to the unavailability of subtype labels in both source and target domains.

Recently, pseudo labels of a target domain have been widely used in UDA \cite{zou2019confidence,liu2020energy}. The pseudo labels are used to estimate target class centers \cite{chen2019progressive}, and the results are enforced to match the source class centers. Contrastive Adaptation Network \cite{kang2019contrastive} is proposed to estimate contrastive domain discrepancy with the target pseudo labels. Considering that the pseudo labels can be noisy, the Gaussian-uniform mixture model is proposed to measure the correctness \cite{gu2020spherical}. In this work, rather than using a sophisticated noisy model, we propose a noise-robust sub-graph scheme with a simple online semi-hard mining method \cite{liu2017adaptive,liu2019hard}. 

 The class-wise conditional alignment~\cite{pan2019transferrable} is also proposed by matching the source and target class centers as in \cite{chen2019progressive}. Moreover, source centers can be regarded as class protocols for classification. Although the source class-wise separation can be enforced by its CE loss \cite{liu2016large}, the center matching does not encourage the compactness of source and target samples, and can lead to sparse target distribution and considerable inner-class variation.      

The center loss \cite{wen2016discriminative} is proposed to encourage the compact distributed representation feature for face identification. Following this line of research, numerous works \cite{liu2017adaptive,liu2018adaptive,xu2020reliable,liu2019hard,liu2019dependency} adapt the center loss to metric learning and optimal transport methods. However, the underlying subtype distribution and shifts are largely ignored. Moreover, the online exploring of the subtype proposed in this work is also closely related to the unsupervised deep clustering \cite{caron2018deep}. Not limited by using $k$-means with the prior knowledge of the subtype numbers, we further propose a scalable sub-graph scheme without the need for the subtype number. Notably, the dynamic memory framework is robust to the pseudo label noise and subtype undersampling.


%% file: 3_Approach.tex
\section{Methodology}

In this work, we consider the UDA task, where we have a source domain $p^s(x,y)$ and a target domain $p^t(x,y)$, and our learning framework has access to a labeled source set $\{(x_i^s,y_i^s)\}$ drawn from $p^s(x,y)$ and an unlabeled target set $\{(x_i^t)\}$ drawn from $p^t(x,y)$. The class label space of $y_i\in\left\{1,2,\cdots,N\right\}$ is shared for both source and target domains. We use $n$ to index $N$ classes. For class $n$, we assume that there are $K_n$ underlying subtypes indexed with $k\in\{1,2,\cdots, K_n\}$. UDA aims to build a good classifier in the target domain $p^t(x,y)$ with the following theorem: \\~

\noindent\textbf{Theorem 1} For a hypothesis $h$ drawn from $\mathcal{H}$, $\epsilon^t(h)\leq$ $\epsilon^s(h)+\frac{1}{2}d_{\mathcal{H}\triangle\mathcal{H}}\{s,t\}+^{\rm{min}}_{h\in\mathcal{H}}[\epsilon^s({h},l_s)+\epsilon^t({h},l_t)]$. \\~

\noindent Here, $\epsilon^s(h)$ and $\epsilon^t(h)$ denote the expected loss with hypothesis $h$ in the source and target domains, respectively. Considering that the disagreement between labeling function $l_s$ and $l_t$, i.e., $^{\rm{min}}_{h\in\mathcal{H}}[\epsilon^s({h},l_s)+\epsilon^t({h},l_t)]$, can be small by optimizing $h$ with the source data \cite{ben2007analysis}, the UDA focuses on minimizing the cross-domain divergence $d_{\mathcal{H}\triangle\mathcal{H}}\{s,t\}$ in the feature space of $f(x_i^s)$ and $f(x_i^t)$.

Instead of aligning $p^s(f(x))$ and $p^t(f(x))$ \cite{kouw2018introduction}, the prototypical networks propose to match the class centroids \cite{pan2019transferrable}. However, the decision boundary of the low-density distributed target sample can be difficult to define, and the inherent subtype structure is underexplored.  
 
On the embedding space, we expect the class separation in the target domain can be achieved when the source domain classes are well-separated, and the class-wise source-target distribution compactness is enforced. Moreover, under the fine-grained subtype local structures and their conditional and label shifts, the subtype-wise tight clustering can be a good alternative to achieve class-wise alignment and compactness. Accordingly, we have the following proposition:\\~

\noindent\textbf{Proposition 1.} The class-wise compactness can be a special case of subtype-wise compactness by assigning the subtype number of this class to 1. \\~

Targeting the conditional and label shifts in both class-level ($p^s(f(x)|y)\neq p^t(f(x)|y), p^s(y)\neq p^t(y))$ and subtype-level ($p^s(f(x)|k)\neq p^t(f(x)|k), p^s(k)\neq p^t(k))$, we propose a novel subtype-aware alignment framework based on an adaptive clustering scheme as shown in Fig. \ref{fig:11}. 
 
\subsection{Class-wise source separation and matching} 

Forcing the separation of classes in the source domain, $p^s(x,y)$, can be achieved by the conventional CE loss \cite{liu2016large}. With the extracted features, we carry out the classification via a remold of the distance to each class cluster center \cite{chen2019progressive,pan2019transferrable}.

For the labeled source data $\{(x_i^s,y_i^s)\}$, we represent the feature distribution of class $n$ with a class centroid $c_n^s=\frac{1}{M_c^s}\sum_{i=1}^{M_c^s}f(x_i^s)$, where ${M_c^s}$ is the involved source sample number. For an input source sample $x_i^s$, we can directly produce a probability histogram with the softmax normalized distance between $x_i^s$ and the centroids $c_n^s$. Specifically, the probability of $x_i^s$ belonging to class $n$ can be formulated as

\begin{equation}
\begin{aligned}
p(y_i^s=n|x_i^s)=\frac{e^{-||f(x_i^s)-c_n^s||_2^2}}{\sum_{n=1}^N e^{-||f(x_i^s)-c_n^s||_2^2}}.\label{eq:1}
\end{aligned}\end{equation} With the one-hot encoding of true class $n$, we define the class-wise CE loss for the source domain samples as $\mathcal{L}_{CE}^{class}=-\text{log} p(y_i^s=n|x_i^s)$.

Following the self-labeling scheme \cite{zou2019confidence,pan2019transferrable}, each target sample $x_i^t$ is assigned with a pseudo class label $\hat{y}_i^t$ to its nearest source centroids, i.e., $\hat{y}_i^t=n$ if $~_{\forall n}^{\text{min}}||f(x_i^t)-c_n^s||_2^2$. With the pseudo class label $\hat{y}_i^t$, we calculate the target domain class-level centroids $c_n^t=\frac{1}{M_c^t}\sum_{i=1}^{M_c^t}f(x_i^t)$, where ${M_c^t}$ is the involved target sample number. We expect the close proximity of $c_n^s$ and $c_n^t$ with $\mathcal{L}^{class}={\frac{1}{N}\sum_{n=1}^N}||c_n^s-c_n^t||_2^2$, which is not sensitive to label shift, since it only chooses the representative centroids of the source and target distribution\footnote{The source \& target center $c_n^{st}=\frac{\sum_{i=1}^{M_c^s+M_c^t}f(x_i^{st})}{M_c^s+M_c^t}$ used in \cite{pan2019transferrable}, and its objective of close proximity of $c_n^s\leftrightarrow c_n^{st}$, or $c_n^t\leftrightarrow c_n^{st}$ are not robust to label shift. Note that $c_n^{st}$ will change if we simply double the involved source/target samples.}. However, neither $\mathcal{L}_{CE}^{class}$ nor $\mathcal{L}^{class}$ considers the inner-class compactness \cite{wen2016discriminative} and the fine-grained subtype structure.

\subsection{Subtype-aware alignment with $K_n$ prior}

If the subtype numbers $K_n$ of class $n$ is known (e.g., 4 subtypes in the CHD disease dataset), we can achieve feature space class-independent clustering with the concise $K$-means, by defining $K$ to be $K_n$.

We denote $K_n$ clustered subtypes with $k\in\{1,2,\cdots, K_n\}$, and calculate the source and target subtype centroids $\mu_k^s$ and $\mu_k^t$, respectively. However, $K$-means does not assign the specific class label to each cluster. To correlate the source and target clusters, we rank the distance of $K_n^2$ subtype centroid pairs and link the smallest rank first. 

Because of the imbalance distribution of subtypes and possible label shift, we assign the subtype centroids of both the source and target samples with $\mu_k^{st}=\frac{\mu_k^s+\mu_k^t}{2}$ instead of averaging all of the source and target samples in subtype $k$. Therefore, each subtype in both source and target domains contributes equally to $\mu_k^{st}$. Then, we enforce all of the samples in subtype $k$ to be close to the subtype centroid $\mu_k^{s,t}$. For the sake of simplicity, we omit the class notation. The subtype compactness objective $\mathcal{L}_k^{sub}$ can be \begin{equation}
\begin{aligned}
   \frac{1}{M_k^s}\sum_{i=1}^{M_k^s}||f(x_i^s)-\mu_k^{st}||_2^2+\frac{1}{M_k^t}\sum_{i=1}^{M_k^t}||f(x_i^t)-\mu_k^{st}||_2^2, \label{eq:2}
\end{aligned}\end{equation} where $M_k^s$ and $M_k^t$ are the numbers of source and target samples in subtype $k$, respectively, to balance the subtype label shift. $\mathcal{L}_k^{sub}$ is traversed for $N$ classes and their $K_n$ subtypes to calculate the sum of normalized subtype compactness loss $\mathcal{L}^{sub}=\frac{1}{N}\sum^N(\frac{1}{K_n}\sum^{K_n}\mathcal{L}_k^{sub})$. Note that an image that does not belong to class $n$ does not belong to any of its $K_n$ subtypes. The class-wise matching and subtype-wise compactness objectives can be aggregated as a hierarchical alignment loss $\frac{1}{N}\sum^N(\alpha\mathcal{L}^{class}+\beta\frac{1}{K_n}\sum^{K_n}\mathcal{L}_k^{sub})$, where $\alpha$ and $\beta$ are the balancing parameters. In the feature space, the learned representations are expected to form $K_n$ compact clusters for class $n$, while each cluster does not need to be far away from one another.

\begin{figure}[t]
\centering
\includegraphics[width=8.8cm]{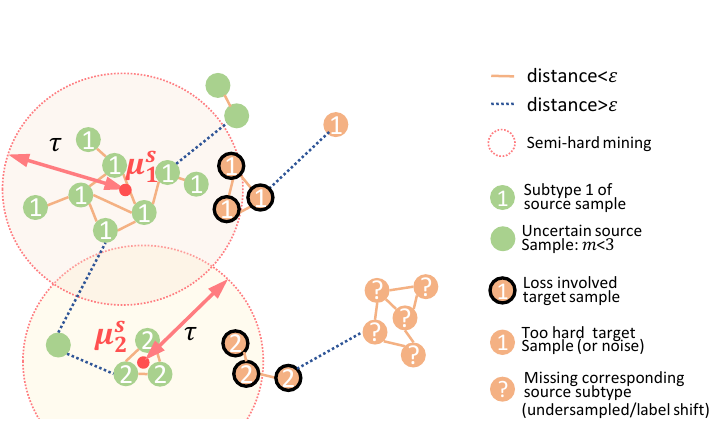}\\ 
\caption{Illustration of the reliability-path based sub-graph construction and alignment with $m=3$. \protect\includegraphics[scale=1.2]{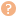} will be assigned to subtype 2, which will then be rejected by  $\tau$.}\label{fig:33} 
\end{figure}

\subsection{Reliability-path based sub-graph construction}

Defining or estimating $K_n$ can be difficult in many applications. Setting $K_n$ of all classes as $N$ hyper-parameters requires costly trails considering the diverse value range of different classes. Note that with a deterministic encoder $f$, the distribution protocol can be similar among different subtypes and classes \cite{fahad2014survey}.

Therefore, we propose constructing the sub-graph with the reliability-path to achieve the online source domain subtype clustering. We assume that similar samples are likely to be distributed closely in the feature space with a deterministic encoder $f$ and form a high-density region \cite{carlucci2019domain}. Given $M^{s}$ samples from the class $n$ in the source domain, there are $(M^{s})^2$ possible edges in a graph. To explore the local structure of the feature space, the two nodes $\{f(x_i^s),f(x_j^s)\}$ are connected by the reliability-path, if $||f(x_i^s)-f(x_j^s)||_2^2\leq\epsilon$. The directly or indirectly linked nodes are combined to form a sub-graph. 

To further eliminate the effect of noise and undersampled subtypes on a batch, we only select the sub-graphs with more than $m$ nodes as the valid subtype clusters.

After exploring the $K_n$ subtypes in the source domain and calculating their centroids $\mu_k^s$, we assign each target sample with the pseudo label of class $n$ to the subtype with the most similar centroid (i.e., $~_{\forall k}^{\rm min}||f(x_i^t)-\mu_k^s||_2^2)$. 

Considering the relatively low confidence or reliability of pseudo target labels \cite{zou2019confidence,gu2020spherical}, we adopt a simple online semi-hard mining scheme to select the target sample in a subtype. The cross-domain margin $\tau$ is used to define a circle at the center of $\mu_k^s$. For the target sample with the initial pseudo subtype label $k$, we choose these samples to distribute within the circle. We note that some target samples may be densely distributed around the circle boundary, and it is not reasonable to cut them apart simply. Therefore, we also resort to the reliability-path to involve the closely distributed neighboring target samples. The sub-graph construction can be robust to missing subtypes in the source or target domains caused by undersampling, since $m$ filters the unreliable source cluster out, and the self-labeling with semi-hard mining of $f(x_i^t)$ rejects the additional subtypes in the sampled target domain. The operation is illustrated in Fig. \ref{fig:33}.

With the reliability-path connected $M_k^s$ source samples and the refined $M_k^t$ target samples in subtype $k$, we calculate $\mu_k^{st}=\frac{\mu_k^s+\mu_k^t}{2}$, and enforce the subtype-wise compactness with $\mathcal{L}_k^{sub}$ as in Eq.~(\ref{eq:2}). 

The online sub-graph construction and alignment have three hyperparameters, including $\epsilon$, $\tau$, and $m$ that are shared for all classes and their subtypes, which can be regarded as the meta-knowledge across clusters. Moreover, we can simplify $\epsilon$ to be the constant 1, and change it to any other positive value results only in the matrices being multiplied by corresponding factors \cite{liu2017adaptive}. The range of $m$ can also be narrow and similar among different subtypes/classes.

\subsection{Optimization and implementation} 

The modern neural networks usually extract a high-dimensional vector, e.g., 4,096 or 2,048-dimensional features, as their representation, thereby demanding high memory and time complexity in subsequent clustering. To remedy this, deep clustering \cite{caron2018deep} proposes to perform dimension reduction via Principal Component Analysis (PCA) for the extracted features from all of the samples in a dataset. PCA, however, is not applicable anymore in our online SubUDA. The feature representations are extracted in different training iterations with different timestamps, which can have incompatible statistics. It is also computationally demanding to carry out PCA for all iterations. Accordingly, a non-linear head-layer with the structure of {fc$\rightarrow$bn$\rightarrow$relu$\rightarrow$dropout$\rightarrow$fc$\rightarrow$relu} is adopted in order to reduce high dimensional features into 256 dimensions. As well, it is simultaneously optimized in each online SubUDA iteration. The non-linear head-layer is eliminated for the subsequent operations, e.g., calculating the L2 distance between features.

In order to prevent the subtype clustering from collapsing to a few subtype groups, \cite{caron2018deep} makes uniform sampling in all of the epochs, which is difficult in our online UDA setting, due to the missing subtype and target class labels. We thus propose a concise approach for SubUDA via re-weighting the loss with $\omega_k\propto\frac{1}{\sqrt{M_k^s+M_k^t}}$, according to the number of samples in the $k$-{th} subtype. Therefore, samples in smaller clusters are accounted more for the loss, which thereby pushes the classification boundary away to incorporate as more samples as possible. The optimization objective can be summarized as \begin{equation}
\begin{aligned}
\mathcal{L}=\mathcal{L}_{CE}^{class}+\frac{1}{N}\sum^N(\alpha\mathcal{L}^{class}+\beta\frac{\omega_k}{K_n}\sum^{K_n}\mathcal{L}_k^{sub}).
\end{aligned}
\end{equation}
 
For the classification in testing, we utilize the centroids of training features as prototypical \cite{pan2019transferrable}.

%% file: 4_Experiments.tex
\begin{figure}[t]
\begin{center}
\includegraphics[width=1\linewidth]{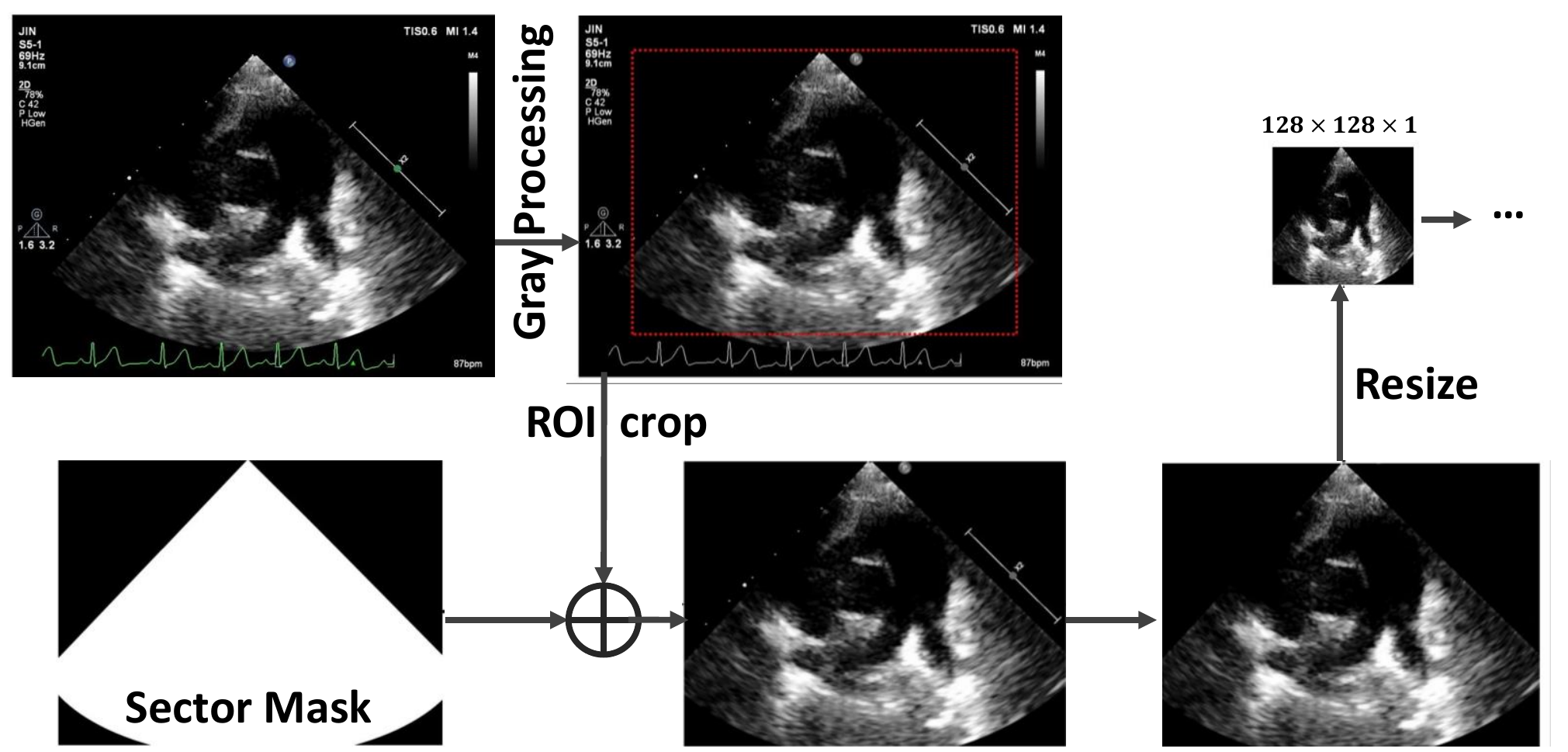}
\end{center} 
\caption{Data pre-processing flowchart.}
\label{fig:2}
\end{figure}

\section{Experiments}

We carried out experiments using CHD data to evaluate the effectiveness of our approach. We implemented our method and comparison methods using PyTorch and set $\alpha=1$, $\lambda=0.5$, and $\beta=0.5$ consistently.

CHD is one of the most common types of birth defect, which usually results in the death of neonates. Therefore, early medical care and treatment can be helpful, which requires an efficient and accurate diagnosis. In clinical practice, clinicians rely on echocardiograms from five heart views. Collected raw data include echocardiogram videos, and a key-frame of each view is usually extracted for the assessment. Specifically, the collected views are from the parasternal long-axis (PSLAX), parasternal short-axis (PSSAX), apical four chambers (A4C), subxiphoid long-axis (SXLAX), and suprasternal long-axis (SSLAX). 

To quantify the effect of subtype structure, four subtypes of CHD, including atrial septal defect (ASD), ventricular septal defect (VSD), patent ductus artery (PDA), and tetralogy of Fallot (TOF), are labeled by two clinicians or intraoperative records. We note that the fine-grained subtype label is not used in training, since large-scale labeling can be costly in clinical practice, while the normal/patient label can be relatively easy to acquire by primary clinicians. Of note, this work focuses on exploring the subtype-aware alignment for the conventional class-wise discriminative model.

Specifically, we evaluated our method on five-view echocardiogram datasets collected from two medical centers. We used 1,608 labeled source subjects (normal/patient) from Beijing Children's Hospital (BCH) and 800 unlabeled target subjects from Beth Israel Deaconess Medical Center (BIDMC). Each dataset consists of echocardiograms from five views that are sufficient for the diagnosis. 

\begin{figure}[t]
\centering
\includegraphics[width=8.5cm]{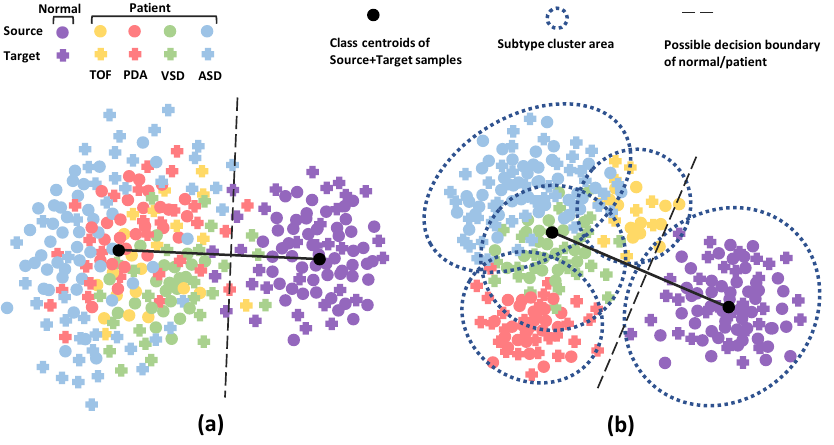}\\ 
\caption{T-SNE visualization of the sampled CHD features with (a) TPN \cite{pan2019transferrable} and (b) our SubUDA. We normalized the distance between class centroids to demonstrate the class separation and inner-class/subtype compactness.}\label{fig:55} 
\end{figure}

 
Abnormal regions are likely to be different from one subtype to another; as such, different subtypes can be easily detected from different views. Considering the large inner-class variation of the patient class, it is reasonable to enforce the subtype-wise compactness. In Fig. \ref{fig:55} left, we can see that the source CHD samples of a subtype also tend to be distributed closely, demonstrating the underlying inner-subtype similarity. With our subtype-wise compactness objective as shown in Fig. \ref{fig:55} right, both the source and target samples are grouped into the high-density region w.r.t. subtypes.
 
\subsection{Data collection of cardiac ultrasound images}

$\bullet$ \textbf{BCH source domain}. A total of 1,608 echocardiogram datasets, including 823 healthy controls, 209 VSD, 276 ASD, 124 TOF, and 176 PDA, were collected using PHILIPS iE 33. The chest of each patient was exposed to the echocardiogram with the supine position. We set our transducer frequency between 3 to 8 MHz. We collected the five standard 2D views, i.e., PSLAX, PSSAX, A4C, SXLAX, and SSLAX.
 
$\bullet$ \textbf{BIDMC target domain}. A total of 800 echocardiogram datasets, including 300 healthy controls, 150 VSD, 150 ASD, 100 TOF, and 100 PDA, were collected. Similarly, the chest of each patient was exposed to the echocardiogram with the supine position. We used PHILIPS EPIQ 7C for echocardiogram imaging and set its transducer frequency from 3 to 8 MHz. We also collected the five standard 2D views, i.e., PSLAX, PSSAX, A4C, SXLAX, and SSLAX.

Notably, discrepancies between these two medical centers include imaging devices (PHILIPS iE 33 vs. EPIQ 7C), patient populations, and clinicians' echocardiogram imaging experience, which introduced domain shifts.

We also extracted the key frame from the video of each view. The heart is a dynamic organ and has different shape within a heart cycle. Thus, we chose a time frame that the clear defects were visible. Specifically, the key frame was corresponding to the isovolumic relaxation phase.


We selected 80\% and 20\% subjects from the target domain for training and testing, respectively. Note that we only selected the subjects with all the key frames from the five views as our testing data. In addition, the key frames shown in training were not used for testing in a subject-independent manner.

\subsection{Pre-processing}

The selected color key frames were first translated to the gray images, since the echocardiogram region has only single-channel 2 gray value information. Then, we cropped the region of interest (ROI) with a redefined sector mask. Since a typical input to convolutional neural networks (CNNs) has the size of 128$\times$128, we resized the masked ROI to the 128$\times$128 image. The five views were stacked following a specific sequence (i.e., SLAX, PSSAX, A4C, SXLAX, and SSLAX) to form our five channels training sample.




 \begin{figure}[t]
\centering
\includegraphics[width=8.5cm]{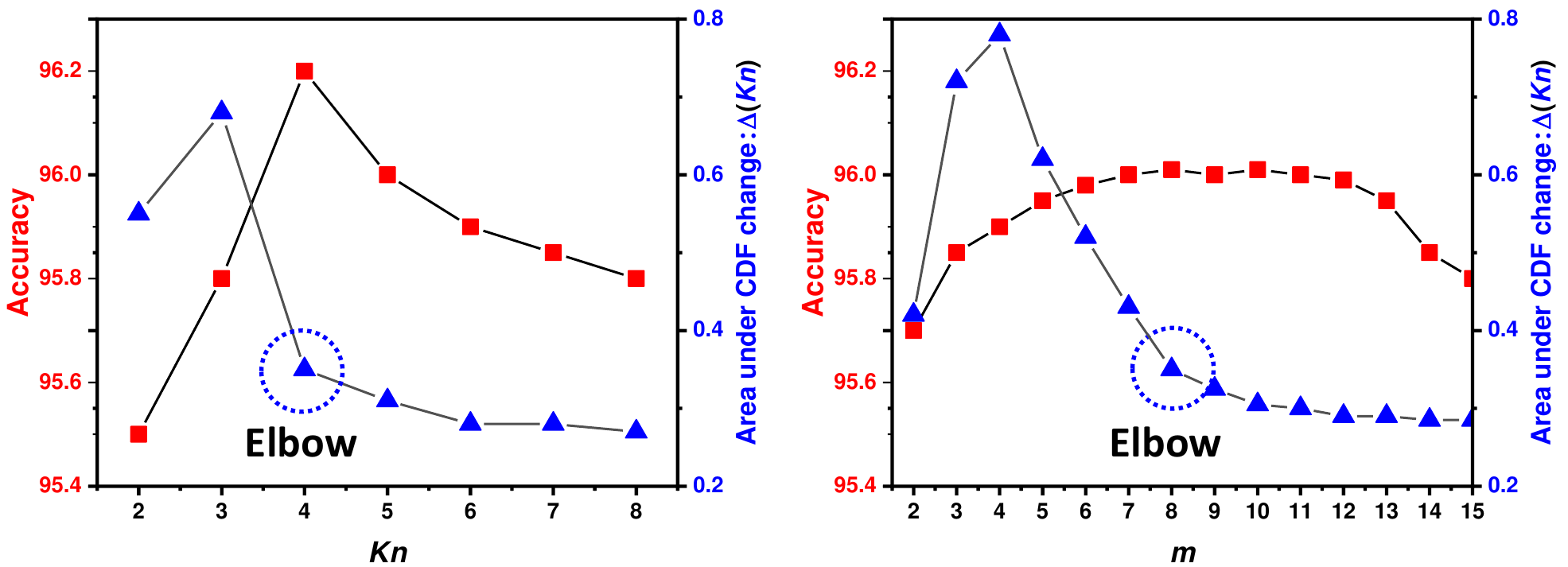}\\ 
\caption{The sensitive analysis of $K_n$ (right) and $m$ of patient class in CHD dataset.}\label{fig:77} 
\end{figure}

\subsection{Backbone network structure for the CHD diagnosis task}

Training a network with massive parameters using a limited number of training data samples usually results in overfitting, which causes a problem in medical image analysis. To alleviate the data constraints in our CHD diagnosis task, we propose to adopt the Depthwise Separable Convolution (DSC) \cite{howard2017mobilenets} as our backbone to reduce the to-be-trained parameters. This strategy was introduced in the MobileNet~\cite{howard2017mobilenets} with a lightweight implementation. Notably, as a comparison, the DSC based CNNs with 1.32M weights performed similarly to AlexNet \cite{simonyan2014very} with 60M weights trained using the ImageNet dataset. 

More specifically, the convolution operation in conventional CNNs is separated to the depth-wise and point-wise stages. The depth-wise stage processes each channel independently, while conventional CNNs process all channels together. Then, the 1$\times1$ convolution is used in the subsequent point-wise stage. We adapt the two-stage convolution for our grayscale image analysis.

\begin{figure}[t]
\begin{center}
\includegraphics[width=1\linewidth]{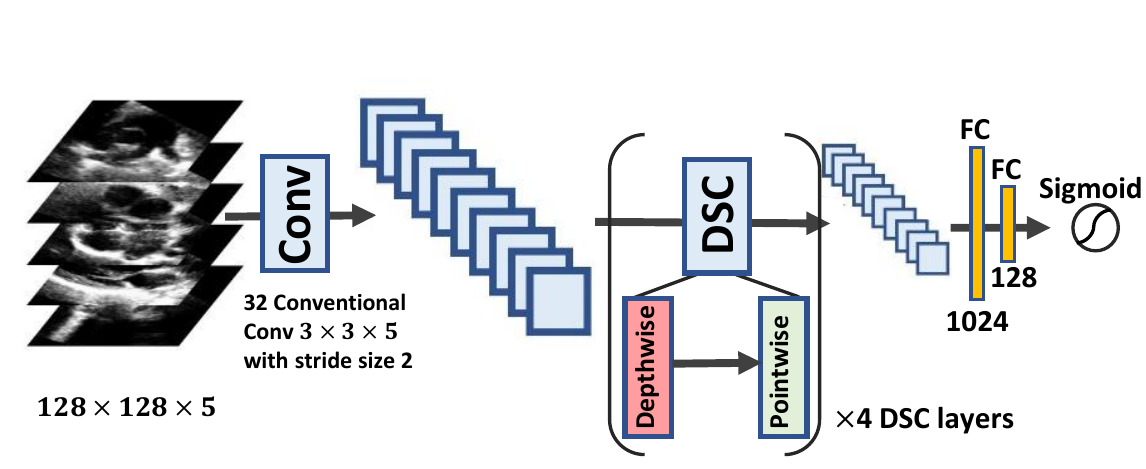}
\end{center} 
\caption{The backbone network via the multichannel convolutional neural networks for five-view echocardiograms analysis.}
\label{fig:4}
\end{figure}

To fuse the information from the five views, we adopted the multi-channel DSC network. Following the MobileNet \cite{howard2017mobilenets}, we configured the first layer as the conventional convolution operation and set the stride as two. We also adopted two fully connected layers with the dimension of 1024 and 128, respectively. The structure of our backbone is shown in Fig. \ref{fig:4}, and detailed in Table 1.





\begin{table}[t]
\caption{{The detailed network construction of the DSC-based five-view framework. We denote the depth-wise convolution with the size $H\times W$ as dw, which is the same for all channels. The point-wise convolution is denoted by pw, which has the size of $1\times 1\times N$. The conventional convolution operation is used in the first convolutional layer.}} 
\centering 
\resizebox{1\columnwidth}{!}{%
\begin{tabular}{l | l | l} 
\hline\hline 
Input Size&Type / Stride & Filter Shape   \\ [0.5ex] 
\hline 

$128\times128\times5$&Conv / s2 & 32 $\times$ of $3\times3\times5$  \\
\hline

$64\times64\times32$&Conv dw / s1& 32 $\times$ $3\times3$ dw \\

$64\times64\times32$& Conv pw / s1& 64 $\times$ of$1\times1\times32$ pw \\
\hline

$64\times64\times64$& Conv dw / s2& 64 $\times$ of$3\times3$ dw \\

$32\times32\times64$& Conv pw / s1& 128 $\times$ of$1\times1\times64$ pw \\
\hline

$32\times32\times128$& Conv dw / s2& 128 $\times$ of $3\times3$ dw  \\

$16\times16\times128$&Conv pw / s1&128 $\times$ of $1\times1\times128$ pw  \\
\hline

$16\times16\times128$&Conv dw / s2&128 $\times$ of $3\times3$ dw \\

$8\times8\times128$& Conv pw / s1&128 $\times$ of $1\times1\times128$ pw \\
\hline

$8\times8\times128$ & Flatten & N/A\\

8192& FC1 & 1024 \\

1024& FC2 & 128 \\
\hline

128& Classifier & Softmax \\
\hline
 
\end{tabular}
\label{table:mobilenet} 
}
\end{table}

In addition to the multi-channel scheme, another feasible choice for multi-view information aggregation is the multi-branch network \cite{lee2016multi}. However, our multi-channel framework has a few strengths over the multi-branch network. First, our multi-channel scheme is able to learn multi-view fusing in all of the layers adaptively, rather than simply concatenating each view in the late layer \cite{lee2016multi}. Second, our multi-channel framework only uses a single forward model, which has much fewer to-be-learned weights. Considering that the echocardiograms from five views can share some similarity, the convolutional filters trained in each view may potentially be useful for one another. Thus, our framework with less to-be-learned weights can efficiently deal with the problem of small dataset size and the requirement of large memory in the implementation.


The class-level classification in our task aims to differentiate healthy controls from patients, which can be formulated as a binary classification problem. We thereby apply the sigmoid unit as our output layer, and adopt the binary CE loss as the supervision signal. We report the accuracy with the threshold of 0.5.

\subsection{Evaluations}

For comparison, we re-implemented the current state-of-the-art methods with the same backbone and experiment setting, where we chose the batch size to 64. The results are shown in Table. \ref{tabel:chd}. Considering the imbalance of normal and patient proportion in the testing set, we also provide the area under the receiver operating characteristic curve (AUC) metric in addition to the accuracy metric.

\begin{table}[]
\centering
\resizebox{1\linewidth}{!}{%
\begin{tabular}{l|cccccc|c}
\hline

Method & Accuracy (\%) $\uparrow$ & AUC $\uparrow$  \\ \hline

Source only& 76.4$\pm$0.12  & 0.721$\pm$0.005 \\

MCD \cite{saito2017maximum}& 88.6$\pm$0.15  & 0.856$\pm$0.003 \\

GTA \cite{sankaranarayanan2018generate}& 90.9$\pm$0.17  & 0.873$\pm$0.005 \\

CRST \cite{zou2019confidence}& 93.2$\pm$0.09  & 0.882$\pm$0.006 \\

TPN \cite{pan2019transferrable}& 93.4$\pm$0.14  & 0.885$\pm$0.004 \\\hline\hline

SubUDA ($K_n=4$)&  {96.2$\pm$0.13}  &  {0.910$\pm$0.003} \\\hline

SubUDA ($K_n=1$)&  {94.7$\pm$0.11}  &  {0.902$\pm$0.004} \\

SubUDA-$\mu^{st}_k$ ($K_n=4$)& 95.4$\pm$0.10  &  0.903$\pm$0.005\\
SubUDA-$\omega_k$ ($K_n=4$)& 96.0$\pm$0.13  &  0.908$\pm$0.004\\
SubUDA-DR ($K_n=4$)&  96.2$\pm$0.11 &  0.911$\pm$0.002\\
\hline

SubUDA-$SG$ ($m=8$)& 96.0$\pm$0.12  & 0.907$\pm$0.004 \\
SubUDA-$SG$-$\tau$ ($m=8$)& 95.5$\pm$0.14  & 0.902$\pm$0.003 \\\hline

\end{tabular}%
}
 
\caption{Experimental results for CHD. $\uparrow$ larger is better.}
 
\label{tabel:chd}
\end{table}

MCD \cite{saito2017maximum} and GTA \cite{sankaranarayanan2018generate} are the typical adversarial training frameworks to align the marginal distribution $p(x)$ at feature level or image level, respectively. The self-training is used to alternatively update the pseudo label of target samples and the network parameters \cite{zou2019confidence}. We note that the compared methods \cite{saito2017maximum,sankaranarayanan2018generate,zou2019confidence,wu2020dual} use the fully-connected classifier after the encoder. The TPN \cite{pan2019transferrable} uses class centroids as a classifier. It proposes to align class centroids of the source and target sample to achieve the conditional alignment w.r.t. $p(x|y)$. Our SubUDA outperformed the state-of-the-art methods w.r.t. both the accuracy and AUC by a large margin, by introducing the subtype-aware constraint. The results indicate that the online subtype compactness can effectively help the classification in the target domain.

\begin{figure}[t]
\centering
\includegraphics[width=8.5cm]{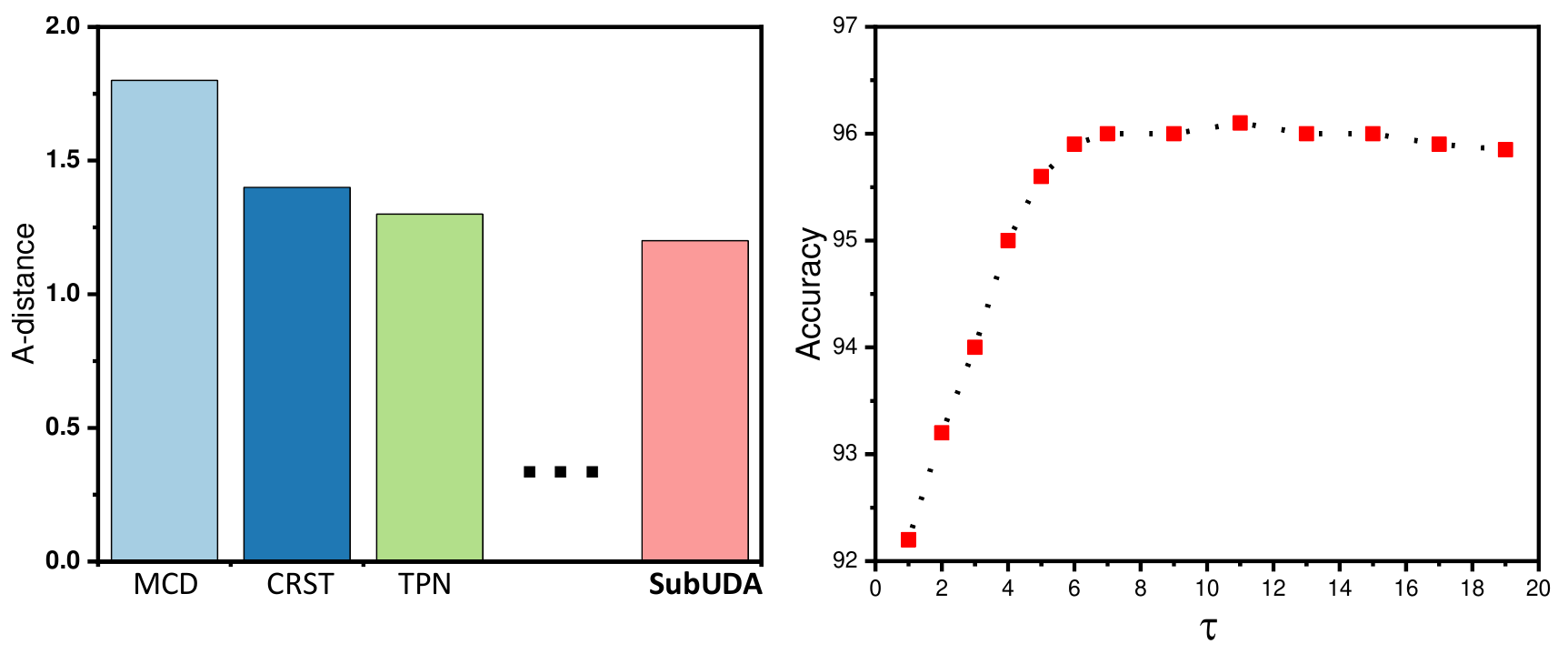}\\ 
\caption{Comparison with the state-of-the-art methods w.r.t. $\mathcal{A}$-distance (left), and the sensitive analysis of $\tau$ (right).}\label{fig:66} 
\end{figure}

The domain adaptation theory suggests proxy $\mathcal{A}$-distance \cite{ben2007analysis} as a measure of cross-domain discrepancy \cite{saito2017asymmetric}. In Fig. \ref{fig:66}, we compare our SubUDA with the other state-of-the-art methods, and the smaller discrepancy has been observed by using the explicit compactness objective in our SubUDA.

For the ablation study, with $K_n=1$, the subtype-wise alignment was reduced to the class-wise compactness. Besides, we used the suffix -DR, -$\omega_k$, and -$\tau$ to denote the subUDA without dimension reduction head, subtype balance weight, and semi-hard target mining, respectively. Furthermore, the suffix -$\mu_k^{st}$ denotes using $\mu_k^{st}=\frac{\sum_{i=1}^{M_k^s+M_k^t}f(x_i^{st})}{M_k^s+M_k^t}$ as the subtype centroid, which is not robust to the subtype label shift. SubUDA-DR took 4$\times$ clustering time, but the improvement was marginal. Therefore, we recommend using the dimension reduction head.


SubUDA-$SG$ in Table \ref{tabel:chd} used the reliability-path based online sub-graph to replace $k$-means. With appropriate $m=8$, the adaptively learned clustering achieved comparable performance to the K-means with $K_n=4$. In Fig. \ref{fig:66} right, we can see that the semi-hard mining scheme in SubUDA-$SG$ is not sensitive to $\tau$ for a large range, since the network can flexibly learn to adjust the ratio of $\epsilon$ and $\tau$ in mapping space \cite{liu2017adaptive}. We note that too strict semi-hard mining (i.e., too small $\tau$) can degenerate our SubUDA to conventional class centroids matching, since no target samples are selected to form a subtype cluster.

In Fig. \ref{fig:77}, we provide a sensitive analysis of hyper-parameter $K_n$ and $m$ for two kinds of online clustering schemes. Consensus clustering can assess the clustering stability \cite{monti2003consensus}, and the optimal clustering is usually achieved in the elbow position of the area under CDF changes, where the CDF is for the consensus matrices\footnote{\url{https://github.com/ZigaSajovic/Consensus_Clustering}}. We can see that the peak of accuracy usually coincides with the best clustering consensus metric, which indicates the good subtype clustering can boost the SubUDA performance. The choice of $K_n=4$ also matches our prior knowledge of the CHD patient subtypes.

Since we have four clear subtypes in this task, using the concise $k$-means can be a straightforward solution. However, Fig. \ref{fig:77} right shows that the accuracy curve can be robust for a relatively large range of $m$, which is promising for the hyperparameter tuning of the case without the prior information of subtype numbers.

%% file: 5_Conclusion.tex
\section{Conclusion}\label{sec:conc}

In this work, we presented a new UDA approach with a more realistic assumption that the subtype-wise conditional and label shifts widely exist, and can be adaptively aligned without the subtype label. We systematically investigate a flexible yet principled solution for the case with and without the prior knowledge of subtype numbers. Rather than the concise $k$-means, we further extend our framework with an online sub-graph scheme using the reliability-path, which can be scalable to many classes and subtypes with a few meta hyperparameters. The effectiveness of explicitly enforcing the subtype-aware compactness has been successfully demonstrated in the CHD transfer task.


\section{Acknowledgement}

We would like to thank anonymous reviewers for their valuable comments.